# A curated, ontology-based, large-scale knowledge graph of artificial intelligence tasks and benchmarks

Kathrin Blagec[1], Adriano Barbosa-Silva[1], Simon Ott[1], Matthias Samwald[1]

**Affiliations**

1. Institute of Artificial Intelligence, Medical University of Vienna, Vienna, Austria.

Corresponding author: Matthias Samwald (matthias.samwald [at] meduniwien.ac.at)

## Abstract

Research in artificial intelligence (AI) is addressing a growing number of tasks through a rapidly growing number of models and methodologies. This makes it difficult to keep track of where novel AI methods are successfully – or still unsuccessfully – applied, how progress is measured, how different advances might synergize with each other, and how future research should be prioritized.

To help address these issues, we created the Intelligence Task Ontology and Knowledge Graph (ITO), a comprehensive, richly structured and manually curated resource on artificial intelligence tasks, benchmark results and performance metrics. The current version of ITO contain 685,560 edges, 1,100 classes representing AI processes and 1,995 properties representing performance metrics.

The goal of ITO is to enable precise and network-based analyses of the global landscape of AI tasks and capabilities. ITO is based on technologies that allow for easy integration and enrichment with external data, automated inference and continuous, collaborative expert curation of underlying ontological models. We make the ITO dataset and a collection of Jupyter notebooks utilising ITO openly available.

## Background & Summary

The past decade led to substantial advances in Artificial intelligence (AI). Increases in computational capacity and the development of versatile machine learning models such as deep convolutional neural networks or the transformer architecture made it possible to tackle a wide variety of tasks that were previously deemed intractable. [1,2]

According to the 2021 AI Index Report published by the AI Index Steering Committee of the Human-Centered AI Institute (Stanford University), the number of AI conference publications increased fourfold between the years 2000 and 2019, while the number of AI journal publications grew from roughly 10,000 in the year 2000 to more than 120,000 in the year 2019. [3] Similarly, AI-related publications on the pre-print server arXiv increased more than sixfold within only five years from 2015 to 2020. [3]

This ever-growing amount of research on AI methods, models, datasets and benchmarks makes it difficult to keep track of where novel AI methods are successfully (or still unsuccessfully) applied, how quickly progress happens, and how different AI capabilities interrelate and synergize. This complexity is amplified by the great variety of AI data modalities (e.g., natural language, images, audio, structured data) and application domains (e.g., web search, biology, medicine, robotics, security, advertising). Furthermore, real-world AI systems and the tasks they address are tightly embedded in complex systems of data creation/consumption, non-AI algorithms and social processes. Understanding AI and its global impact requires the creation of rich models that integrate data from these adjacent knowledge domains.

To help address these issues, we introduce the Intelligence Task Ontology and Knowledge Graph (ITO), a comprehensive, richly structured and manually curated data resource on artificial intelligence tasks, benchmark results and performance metrics. ITO is realized as an ontology-backed knowledge graph [4] based on standards minted by the World Wide Web Consortium (W3C). Data are represented through the Resource Description Framework (RDF) [5] and Web Ontology Language (OWL) [6] standards and can be queried through the SPARQL graph query language [7]. These standards have a long history of application in other domains requiring complex knowledge representation and integration, such as biomedical research [8–10].

The following desiderata guided the creation of ITO:

- Manual curation of AI task classification hierarchies and performance metrics, enabling more precise analyses.
- Representing data as a graph, facilitating network-based query and analysis.
- Allowing for easy integration and enrichment with external data, as well as simple extensibility for modelling related knowledge domains from other domains.
- Allowing for automated deductive inference and automated knowledge base consistency checking.
- Allowing for ongoing, collaborative expert curation of underlying ontological models.

ITO allows for capturing rich relationships between AI processes, models, datasets, input and output data types (e.g., text, video, audio) metrics, performance results and bioinformatics processes. This enables a more in-depth tracking of progress over time, such as analysing how progress trajectories on various classes of tasks compare to each other across different dimensions.

In its current version (v1.0), ITO encompasses more than 50,000 data entities and 9,000 classes.

**Exemplary use cases**

The primary aim of ITO is to enable **'meta-research'** concerned with studying scientific research itself in terms of its methods, reporting, evaluation and other aspects to increase the quality of scientific research. [11] The value of ITO for meta-research can be exemplified based on two recent studies utilizing the resource.

As a first step in creating insights from ITO, our group analysed the prevalence of performance metrics currently used to measure progress in AI using data on more than 30,000 performance results across more than 2,000 distinct benchmark datasets. To increase data quality and enable a thorough analysis, we conducted extensive manual curation and annotation of the raw data as part of integrating it into the ontology. [12]

In another recent study, we explored and mapped AI capability gains over time across 16 main research areas (e.g., computer vision, natural language processing, graph processing), further breaking down capability gain by sub-processes described in the curated AI process hierarchy (Barbosa da Silva et al., manuscript in preparation). This analysis made use of both the curated AI process class structure, as well as the cleaned and normalized performance metrics data.

Besides enabling meta-research, the ontology of ITO can be **utilized as a taxonomic resource for annotating and organizing** information in the AI domain. For example, in recent work, our group conducted a systematic review of literature and online resources to create a catalogue of AI datasets and benchmarks for medical decision making. Identified datasets and benchmarks were manually annotated for meta-information, such as targeted tasks, data types and evaluation metrics. This manual curation process was greatly simplified by using the taxonomic structures provided by ITO together with the OntoMaton ontology annotation widget. [13,14]

Finally, the ITO knowledge network can serve as a practice-focused resource that allows developers to **find, compare and combine AI models to address complex use-cases** for certain defined tasks, data types and application domains.

## Methods

Benchmark result and initial task description data was drawn from the 'Papers with code' (PWC, [https://paperswithcode.com](https://paperswithcode.com)) repository. PWC is the largest repository of AI benchmark data currently available. It is a web-based open platform that contains information on more than 5,000 benchmarks and 50,000 publications. PWC data was collected by combining automatic extraction from arXiv submissions and manual crowd-sourced annotation of benchmark results.

To create the initial version of ITO, PWC data was imported and converted to RDF/OWL with a Python script. After initial data import from PWC, ITO underwent extensive further manual curation. Collaborative manual curation was done using WebProtégé. [15] Curation was conducted by two experts in the AI / machine learning domain (KB, MS) over the course of several months. In this process, tasks and benchmarks were systematised and mapped to so-called 'AI processes'.

The scripts and overall workflow created allow for repeated updating and incremental curation of data over time, so that ITO can be kept up-to-date as benchmark result data in Papers With Code (and potentially other data repositories) keeps evolving.

### Process-centric modeling of AI tasks

In ITO, an 'AI process' is defined as a process that can be carried out (or can partially be carried out) by an AI system. These AI processes are organised into 16 major parent classes, e.g, 'Natural Language Processing', 'Vision process' or 'Audio process', and further into a hierarchy of subclasses informed by the common terminology of the respective research field. For example, the branch of the class 'Natural language processing' was informed both by terminology and taxonomies used in the fields of linguistics and machine learning.

In the current version of ITO, AI processes are represented through a simple, asserted polyhierarchy. Since some AI tasks are cross-modal, child classes can have more than one parent class (multiple inheritance). For example, the process 'Image question answering', which is concerned with answering questions based on the semantic content of an image, has both 'Natural language processing' and 'Vision process' as its superclasses. This application-centric modelling approach was deemed appropriate for currently targeted use cases of ITO, but we plan to utilize more elaborate ontological modeling in future work (e.g., breaking up the process class hierarchy into several complementary axes and making greater use of logical class definitions).

Instances of AI models, e.g. 'BERT' or convolutional neural networks, and benchmark datasets, e.g., 'ImageNet', are modelled via the 'Data' and 'Software' branches of ITO.

**Modeling of benchmarks**

Individual benchmark results were captured as instances of the respective benchmark class and connected to the respective dataset via a 'has_input' annotation. Performance measures (e.g., F1 score) were modelled through a hierarchy of data properties.

Similar to the AI process classes, performance measure properties underwent extensive manual curation. The original data obtained from PWC contained more than 800 different strings representing metric names that were used by human annotators on the PWC platform to add model performance results. Based on this raw list containing more than 60 naming variations of the same metric in some cases, we created a canonical hierarchy of performance measures, and mapped the strings accordingly. To select the canonical names for the metrics, a preference was given to Wikipedia article titles whenever sensible. Additional details and complexities of this process are described in previous work. [12]

Figure 1 shows an example of a benchmark result achieved by a specific *model* on a specific *dataset* for a specific *AI process* embedded in ITO.

Figure 1: Example of a benchmark result for a specific model ('DeBERTa-1.5B') on a specific dataset ('Words in Context', Word sense disambiguation) embedded in ITO.

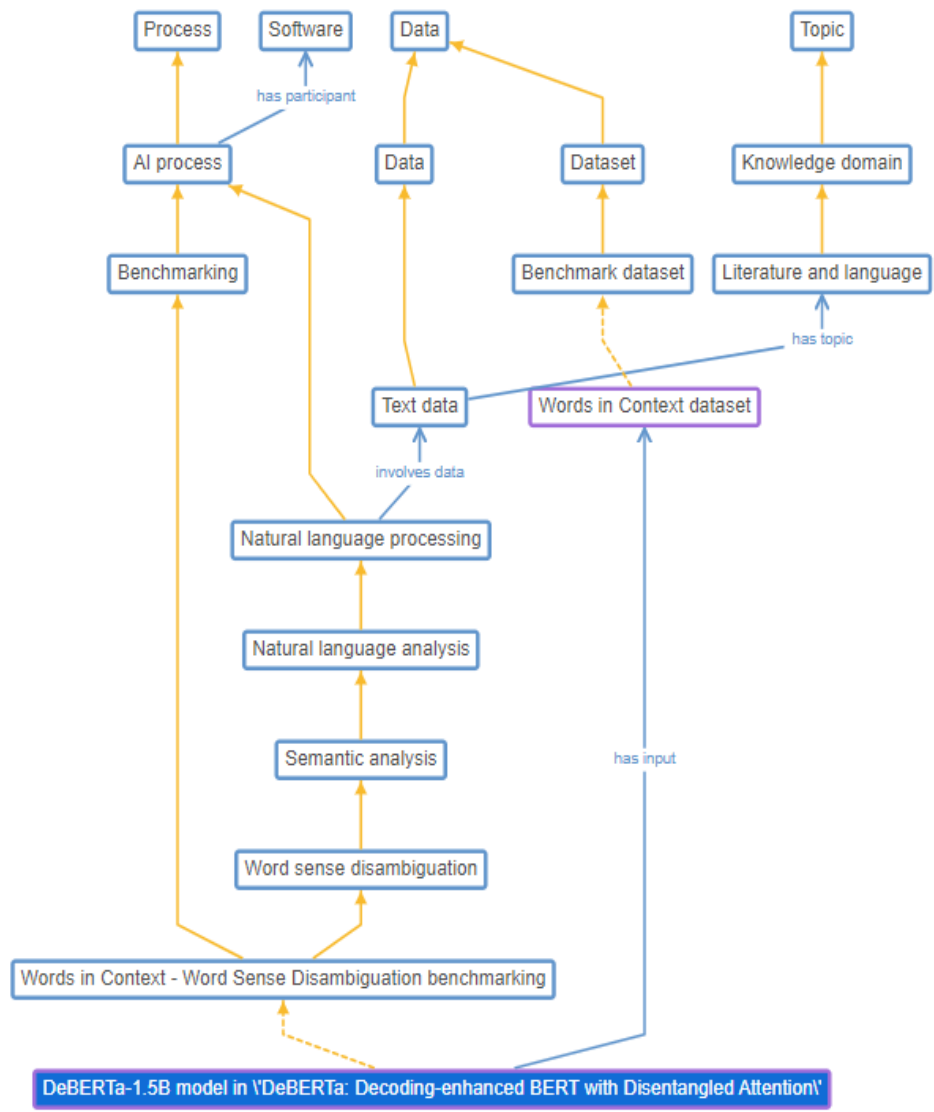

# Data Records

The ITO dataset is made available as a single OWL (Web Ontology Language) file (see Section 'Availability and formats'). The current version of ITO (v1.0) encompasses more than 50,000 individuals across more than 9,000 classes. Additional basic metrics are shown in Table 1.

**Table 1: Basic ontology metrics of ITO (v1.0)**

| Entities | Count |
| --- | --- |
| Total triples (i.e. edges in the graph) | 685,560 |
| Classes (total) | 9,037 |
| Classes (AI process classes) | 1,100 |
| Individuals | 50,826 |
| Object properties | 16 |
| Data properties (i.e. AI performance measures) | 1,995 |
| Annotation properties | 32 |
| Maximum depth | 11 |

In total, ITO captures more than 26,000 benchmark results across more than 3,633 benchmark datasets covering the years 2000 to 2021 (see Table 2 and Figure 2).

**Table 2: Content metrics (v1.0)**

|  | Count |
| --- | --- |
| Total number of papers covered | 7,649 |
| Time span of publications covered | 2000-8/2021 |
| Total number of benchmark results | 26,495 |
| Total number of benchmark datasets | 3,633 |

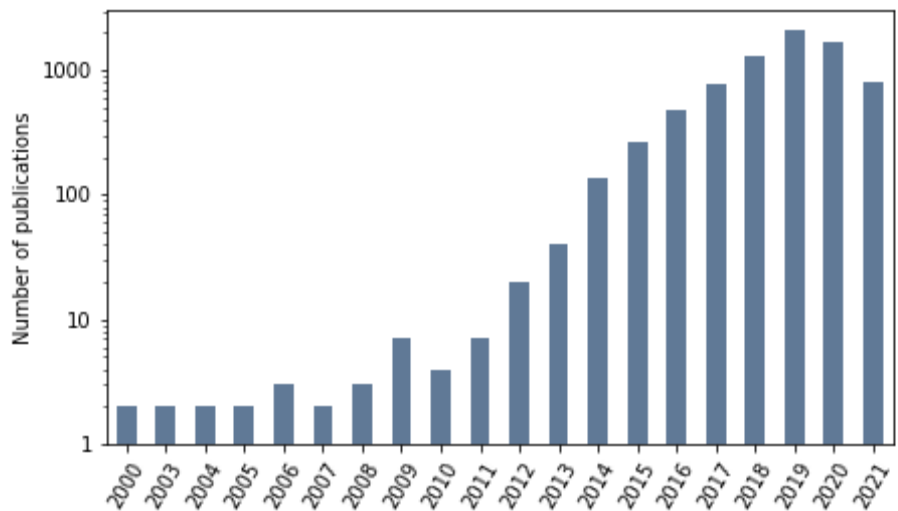

**Figure 2: Number of papers covered by ITO per year.** The y-axis is scaled logarithmically. Publications of the year 2021 are covered until the latest import in August 2021.

Figure 3 shows the 16 parent process classes, e.g, 'Natural Language Processing', 'Computer vision' or 'Audio process' that are used to map AI processes in ITO together with the number of distinct benchmarks and benchmark results per process. An excerpt of the curated performance measure hierarchy is displayed in Figure 4.

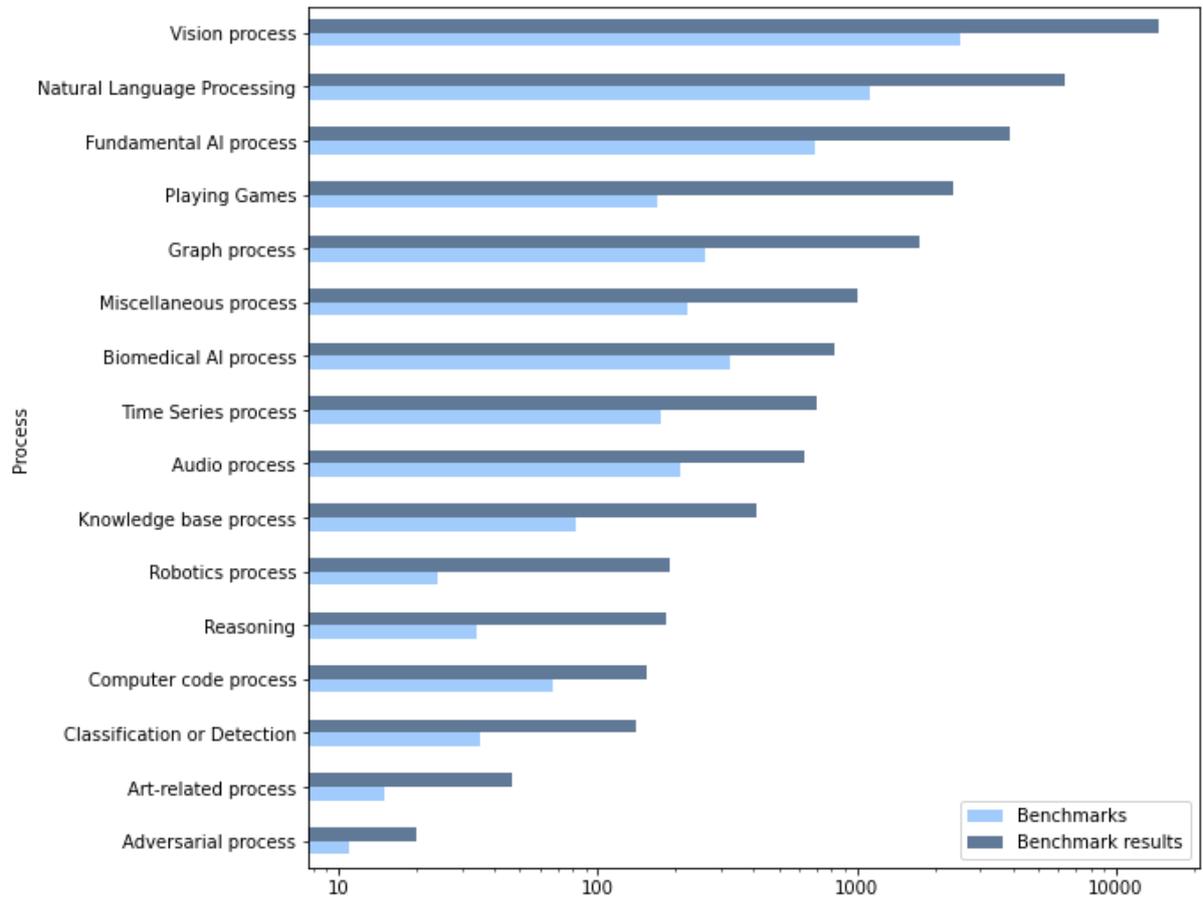

**Figure 3: Number of distinct benchmarks and benchmark results per 'AI process' class.** The x-axis is scaled logarithmically.

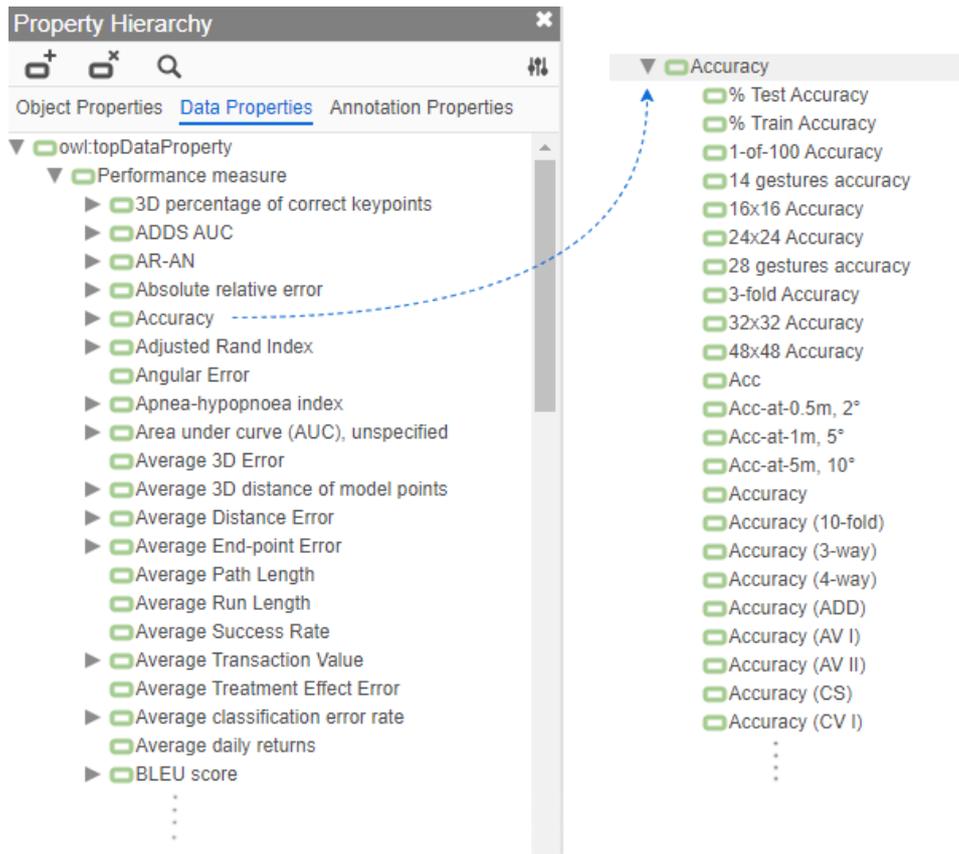

**Figure 4: Performance measures property hierarchy.** The left side of the image shows an excerpt of the list of performance metric properties; the right side shows an excerpt of the list of subclasses for the parent class 'accuracy'. [12]

## Availability and formats

The ITO model is available as an OWL (Web Ontology Language) file. The latest version of ITO is available on both Github (https://github.com/OpenBioLink/ITO) and Bioportal (https://bioportal.bioontology.org/ontologies/ITO). The ontology file is distributed under a CC-BY-SA license.

## Technical Validation

Validation and evaluation of a knowledge graph and/or ontology aims to assess whether the resource adequately and accurately covers the domain it intends to model, and whether it enables an efficient execution of the tasks it was designed for.

Commonly used criteria to evaluate ontologies based on these aspects include *accuracy*, *clarity*, *completeness*, *conciseness*, *adaptability*, *computational efficiency* and *consistency*. [16]

*Accuracy* indicates whether the definitions and descriptions of elements in an ontology are correct. *Clarity* measures whether the ontology's elements are clearly defined and labeled, and understandable for the user. Achieving high accuracy and clarity was ensured in ITO through an extensive manual curation period lasting several months.

The criterion of *completeness* is concerned with whether the domain to be modelled is adequately covered by the ontology, while *conciseness* indicates to which extent the ontology covers only elements relevant to the domain. Both criteria are ensured in ITO through the bottom-up development approach that makes use of existing data (i.e., benchmarks extracted from preprint servers) and concepts relevant to the domain of AI processes instead of a top-down approach that starts with a blank slate. Relying on existing data sources, such as the PWC database that combines automated extraction of benchmarks from papers on preprint servers and crowd-sourced annotation by several thousands of contributors enables high domain coverage.

*Adaptability* is concerned with whether the ontology meets the requirements defined by the range of use cases for which it was built. The practical usability of ITO for its intended applications has been validated within two recently conducted studies (Barbosa da Silva et al., manuscript in preparation). [12]

*Computational efficiency* indicates whether the ontology's anticipated tasks can be fulfilled within reasonable time and performance frames using the available tools. Even complex queries related to the use cases described above can be executed within a few seconds on standard hardware when using the high-performance Blazegraph graph database.

Finally, *consistency* requires the ontology to be free from any contradictions. Internal consistency was checked using Protégé v5.5.0 and the elk 0.4.3 reasoner. [17,18]

Furthermore, common pitfalls in ontology design and creation have been described, which, for example, include the creation of unconnected ontology elements, missing human readable annotations or cycles in class hierarchies. [19–22] ITO was checked for these, and any issues were resolved.

To ensure content validity and keeping up with the fast-paced developments in the field of AI, newly available data will be periodically imported. Further, the underlying ontological model will be subject to continuous refinement, and future developments will also focus on creating mappings between ITO and other thematically relevant ontologies and knowledge graphs, such as the Artificial Intelligence Knowledge Graph (AI-KG) that contains a large collection of research statements mined from AI manuscripts [23].

**Other data sources**

Besides PWC, we also investigated some other projects aiming to track global AI tasks, benchmarks and state-of-the-art results have been initiated in recent years as potential data sources. Among these, the *Aicollaboratory* [24] and *State of the art AI* (https://www.stateoftheart.ai/) stood out as the most comprehensive and advanced resources.

'AIcollaboratory' is a data-driven framework enabling the exploration of progress in AI. It is based on data from annotated AI papers and open data from, e.g., PWC, AI metrics and OpenML. Similar to the projects described above, benchmark results are organized hierarchically and can be compared per task. In addition, the platform provides summary diagrams that combine all benchmark results per top-level task class, e.g., 'Natural language processing' and display progress over time. We found that relevant data in AIcollaboratory were already covered by PWC, and that the project did not seem to be actively maintained at the moment.

'State of the art AI' collects AI tasks and datasets, models and papers building on data from PWC, arXiv, DistillPub and others. Similar to PWC, it organises AI tasks, allows for a comparison of results per task, and makes them available on a web-based platform. However, data are not available for download at the time of this writing, and the of relevant data were already covered by PWC.

## Usage Notes

A wide variety of frameworks for OWL, RDF and the SPARQL graph query language can be used to access and query the ontology. Our recommendations for efficient processing include the graph database Blazegraph (https://blazegraph.com) for both simple and complex queries requiring high performance, and the Owlready2 Python library (https://pypi.org/project/Owlready2/) for simple queries and OWL reasoning.

Example Jupyter notebooks for querying the ontology using the libraries mentioned above can be found in the associated Github repository (https://github.com/OpenBioLink/ITO) in the folder 'notebooks' (e.g., 'descriptive_statistics_v1.0' and 'trajectories_notebooks').

To view and edit the ontology, the Protégé ontology editor (https://protege.stanford.edu/) can be used. [17] Furthermore, the class structure of ITO can be browsed online via BioPortal (https://bioportal.bioontology.org/ontologies/ITO).

## Code Availability

Code to generate the summary statistics are available from the Github repository in the folder 'notebooks': https://github.com/OpenBioLink/ITO

## Acknowledgements

The research leading to these results has received funding from netidee under grant agreement 5158 ('Web of AI') and the European Community's Horizon 2020 Programme under grant agreement No. 668353 ('Ubiquitous Pharmacogenomics').

The authors thank the team of Papers with Code for making their data available.

This work was conducted using the Protégé resource, which is supported by grant GM10331601 from the National Institute of General Medical Sciences of the United States National Institutes of Health.

## Authors contributions

M.S. conceived the project. M.S. and K.B. performed data curation and ontology engineering. M.S., K.B. and A.B.S. wrote code to process and analyse the data, KB prepared the figures, K.B. and M.S. drafted the manuscript, A.B.S. and S.O. reviewed the manuscript. M.S. acquired funding and supervised the project. All authors have read and approved the final manuscript.

## Competing interest

The authors declare that there are no conflicts of interest.